\documentclass[10pt,twocolumn,letterpaper]{article}

\usepackage[pagenumbers]{cvpr} 

\definecolor{cvprblue}{rgb}{0.21,0.49,0.74}
\usepackage{multirow}
\usepackage{arydshln}
\usepackage[pagebackref,breaklinks,colorlinks,allcolors=cvprblue]{hyperref}

\definecolor{Lavender}{RGB}{230, 230, 250}
\definecolor{indigo}{rgb}{0.294, 0.0, 0.51}

\def\paperID{7721} 
\def\confName{CVPR}
\def\confYear{2025}

\newcommand{\fref}[1]{Figure~\ref{#1}}

\newcommand{\sref}[1]{Section~\ref{#1}}
\newcommand{\tref}[1]{Table~\ref{#1}}

\title{AirRoom: Objects Matter in Room Reidentification}

\author{
Runmao Yao \quad
Yi Du \quad
Zhuoqun Chen \quad
Haoze Zheng \quad
Chen Wang \\
Spatial AI \& Robotics (SAIR) Lab, University at Buffalo \\
{\tt\small \{yaorunmao, zhz19231211\}@gmail.com, \{yid, chenw\}@sairlab.org, zhc057@ucsd.edu}
}

\begin{document}
\maketitle
\begin{abstract}
Room reidentification (ReID) is a challenging yet essential task with numerous applications in fields such as augmented reality (AR) and homecare robotics. Existing visual place recognition (VPR) methods, which typically rely on global descriptors or aggregate local features, often struggle in cluttered indoor environments densely populated with man-made objects. These methods tend to overlook the crucial role of object-oriented information. To address this, we propose AirRoom, an object-aware pipeline that integrates multi-level object-oriented information—from global context to object patches, object segmentation, and keypoints—utilizing a coarse-to-fine retrieval approach. Extensive experiments on four newly constructed datasets—MPReID, HMReID, GibsonReID, and ReplicaReID—demonstrate that AirRoom outperforms state-of-the-art (SOTA) models across nearly all evaluation metrics, with improvements ranging from 6\% to 80\%. Moreover, AirRoom exhibits significant flexibility, allowing various modules within the pipeline to be substituted with different alternatives without compromising overall performance. It also shows robust and consistent performance under diverse viewpoint variations. Project website: \href{https://sairlab.org/airroom/}{\textcolor[rgb]{0.3,0.5,1}{https://sairlab.org/airroom/}}.

\end{abstract}    
\vspace{-5pt}
\section{Introduction}
\label{sec:intro}
\vspace{-2pt}

With the rapid development of spatial computing, room reidentification (ReID) has become a key area of interest, enabling advancements in applications like augmented reality (AR) \cite{schult2023controlroom3droomgenerationusing} and homecare robotics \cite{sarch2022tideetidyingnovelrooms}. It plays a crucial role in enhancing user experiences across various scenarios. 
For instance, on devices like the Apple Vision Pro, accurate room ReID enables smooth transitions between virtual and real-world elements. 
Similarly, in AR-guided museum tours, precisely identifying a user’s position within specific rooms is essential for delivering location-sensitive content.

\begin{figure}[ht]
    \centering
    \includegraphics[width=\columnwidth]{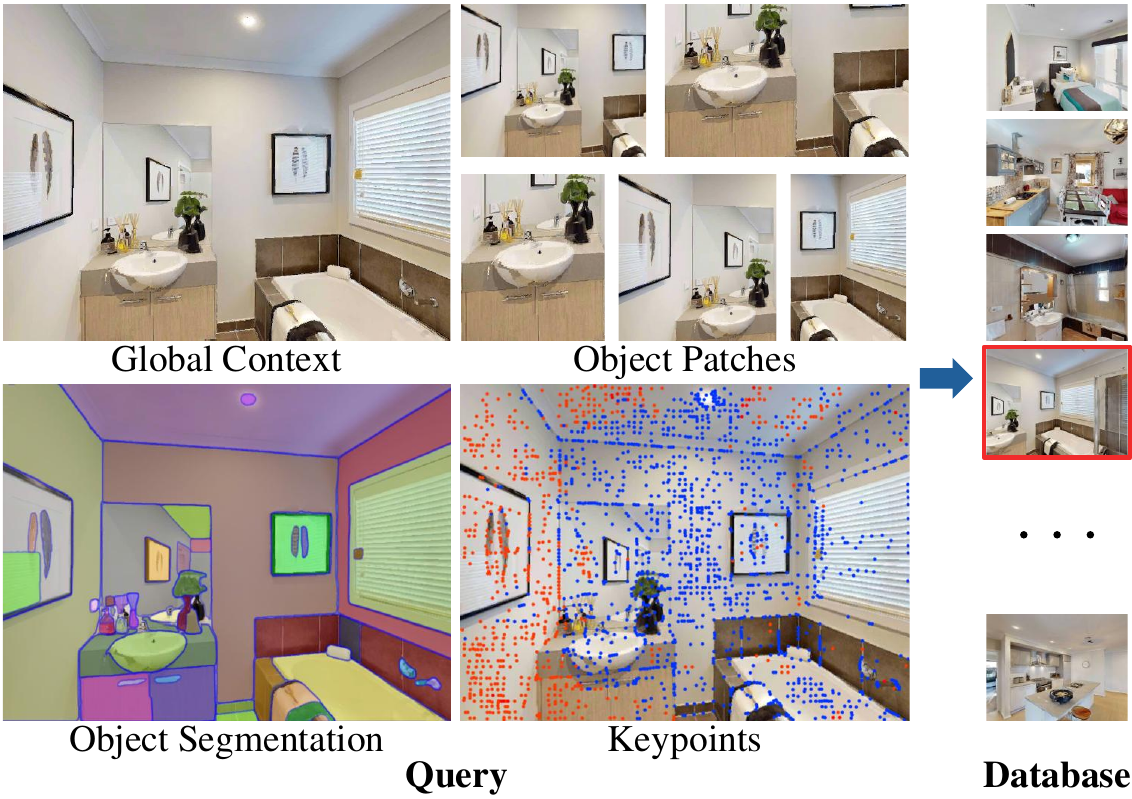}
    \vspace{-20pt}
    \caption{AirRoom leverages multi-level, object-oriented features, including global context, object patches, object segmentation, and keypoints, to perform coarse-to-fine room reidentification.}
    \vspace{-20pt}
    \label{fig:example_image}
\end{figure}

Unlike outdoor environments, where visual place recognition (VPR) methods have matured and perform reliably \cite{arandjelović2016netvladcnnarchitectureweakly, hausler2021patchnetvladmultiscalefusionlocallyglobal, keetha2023anylocuniversalvisualplace}, indoor room ReID remains a challenging problem. A primary reason for this difficulty is the cluttered nature of indoor scenes, which are often densely packed with man-made objects \cite{xu2023clusvprefficientvisualplace}. These densely distributed objects often pose significant challenges to existing methods, which were originally designed for city-style and distinct structures \cite{7339473}. Consequently, these methods struggle to fully capture the intricate details and varied spatial layouts of indoor environments.
For instance, foundation models like DINO \cite{caron2021emergingpropertiesselfsupervisedvision} and DINOv2 \cite{oquab2024dinov2learningrobustvisual} can generate global descriptors that capture broad scene-level features. However, these descriptors may struggle in semantically similar environments, such as adjacent rooms with similar layouts or decorations, where distinguishable features are minimal \cite{cai2022patchnetvladlearnedpatchdescriptor}. In contrast, methods like Patch-NetVLAD \cite{hausler2021patchnetvladmultiscalefusionlocallyglobal}, AirLoc \cite{aryan2023airlocobjectbasedindoorrelocalization} and AnyLoc \cite{keetha2023anylocuniversalvisualplace} create a global descriptor by aggregating local features, which can enhance discriminative power. Yet, in indoor settings densely populated with similar and repetitive objects, these approaches may still face difficulties in distinguishing between highly similar features, reducing their effectiveness in such contexts \cite{sattler2019understandinglimitationscnnbasedabsolute}.

Additionally, different from room categorization \cite{lee2017roomnetendtoendroomlayout}, which relies on identifying object types to classify spaces into semantic categories, 
room ReID requires accurately retrieving the same room instance from a reference database based on a given query image. 
For instance, reidentifying a particular kitchen demands a combination of global functional contexts and fine-grained matching of specific object attributes. Moreover, room ReID must handle viewpoint variations, which necessitates tolerance for partial mismatches in object arrangement and appearance. These requirements often result in the failure of algorithms based solely on object categorization, as they lack the precision needed to reidentify unique room instances accurately \cite{Snderhauf2015PlaceRW}.

This raises an important question: \textit{“What kinds of object attributes are truly essential for room ReID?”} To address this, we conduct the first comprehensive study exploring multi-level object-oriented information and its impact on room ReID.
As shown in \fref{fig:example_image}, our experiments show that all four levels of object-oriented information, \ie, global context, object patches, object segmentation, and keypoints, are essential.
Specifically, we find that each level plays a unique role in room ReID. Global context, such as the combination of objects like a couch and television, conveys essential semantic information for categorizing a room as a living room. Object patches provide finer details, enabling differentiation within a room, such as distinguishing a bedside table in a bedroom from a desk in a workspace. Object segmentation offers further granularity by isolating individual items, like separating a dining table from surrounding chairs to clarify the room layout. Finally, keypoints on objects, such as handles on a dresser, enhance room ReID by filtering out visually similar furniture in other rooms. Moreover, integrating multi-level object-oriented information adds robustness to viewpoint variations.

Based on these observations, we propose AirRoom, a simple yet highly effective room reidentification (ReID) system consisting of three stages: Global, Local, and Fine-Grained. In the Global stage, a Global Feature Extractor is used to capture global context features, which are then employed to coarsely select five functionally similar candidate rooms. In the Local stage, instance segmentation is applied to identify individual objects, followed by the Receptive Field Expander to extract object patches. An Object Feature Extractor is then used to obtain both object and patch features, which are utilized in Object-Aware Scoring to narrow the selection down to two candidate rooms. Finally, in the Fine-Grained stage, feature matching is employed to precisely identify the final room.

In summary, our contributions include:

\begin{itemize}[leftmargin=2em]
    \item We introduce AirRoom, an object-aware room ReID pipeline with two novel modules: the Receptive Field Expander and Object-Aware Scoring, effectively leveraging multi-level object-oriented information to overcome the limitations observed in previous methods.
    \item We have curated four comprehensive room reidentification datasets—MPReID, HMReID, GibsonReID, and ReplicaReID—providing diverse benchmarks for evaluating room reidentification methods.
    \item Extensive experiments demonstrate that AirRoom outperforms SOTAs, maintaining robust and reliable performance even under significant viewpoint variations.
\end{itemize}

\section{Related Work}
\label{sec:related_work}

In this section, we review areas mostly related to our work, \ie, image retrieval and visual place recognition.

\subsection{Image Retrieval}

Image retrieval is a fundamental and well-established task in computer vision that involves searching for images similar to a given query within a large database.
The process of image retrieval typically consists of two stages: global retrieval and re-ranking. In the first stage, a global descriptor that aggregates local features is used to retrieve $k$ candidates from a large database. This is followed by spatial verification through local feature matching to re-rank these $k$ candidates. Early research relied on handcrafted features \cite{Lowe2004DistinctiveIF, BAY2008346}, while current methods utilize deep networks to learn informative representations \cite{cao2020unifyingdeeplocalglobal, radenović2018finetuningcnnimageretrieval}.

Most image retrieval methods focus on selecting diverse relevant images to help users discover options that align with their interests or needs in real-world applications \cite{Wan2014DeepLF}. Although these methods are effective in retrieving similar images, they often lack the emphasis on distinguishing between categories or achieving precise ReID \cite{10.1145/1348246.1348248}.
In \mbox{contrast}, our approach prioritizes achieving accurate ReID. Following a ``global retrieval and re-ranking" pipeline, we first use global context features to identify the top five room candidates. Our object-aware mechanism then refines the search in a coarse-to-fine manner, progressively distinguishing among candidates until the most similar room is \mbox{identified}, yielding accurate results.

\begin{figure*}[ht]
    \centering
    \includegraphics[width=\textwidth]{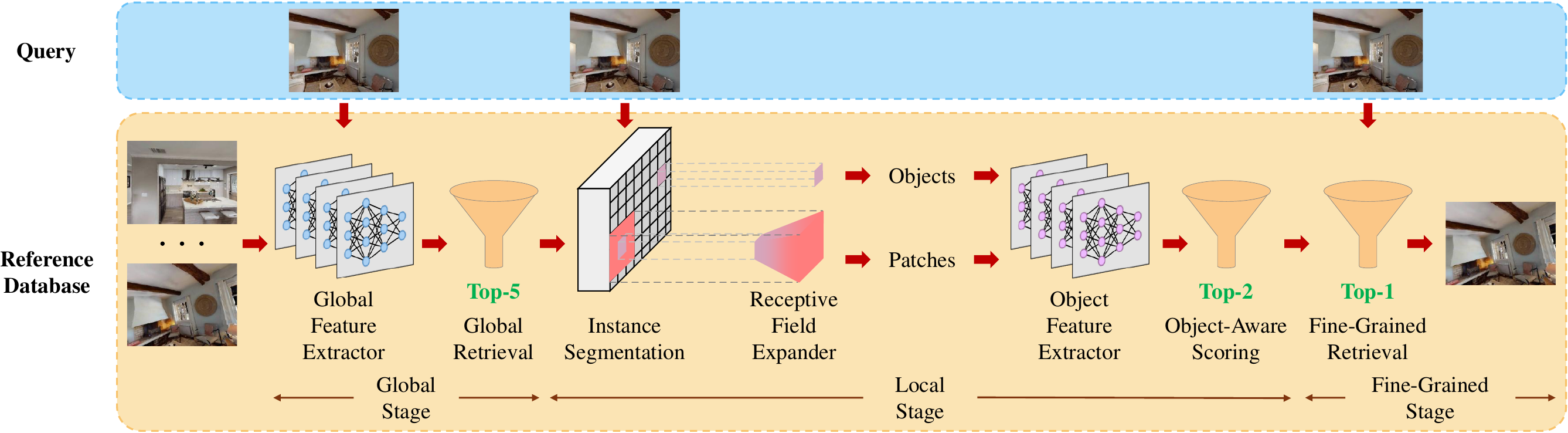}
    \vspace{-16pt}
    \caption{\textbf{The AirRoom coarse-to-fine pipeline}. The pipeline begins with the Global Feature Extractor, which captures global context features to retrieve the top-5 reference images. Instance segmentation then generates object masks, followed by the Receptive Field Expander, which extracts object patches. The Object Feature Extractor processes both object and patch features. The Object-Aware Scoring module narrows the selection to the top-2 candidates, and Fine-Grained Retrieval identifies the most suitable reference image.}
    \vspace{-15pt}
    \label{fig:pipeline}
\end{figure*}

\subsection{Visual Place Recognition}

Visual place recognition (VPR) is often framed as a special image retrieval problem, aiming to match a view of a location with an image of the same place taken under different conditions.
Previous methods fall into two categories: those that directly use global descriptors and those that aggregate local features into a global descriptor. Earlier approaches that relied on global descriptors primarily used CNN-based backbones, such as ResNet \cite{he2015deepresiduallearningimage}, to generate these descriptors. More recent methods, however, leverage foundation models like DINOv2 \cite{oquab2024dinov2learningrobustvisual} for enhanced feature representation. In the aggregation category, early techniques employed handcrafted features like SIFT \cite{Lowe2004DistinctiveIF}, SURF \cite{10.1007/11744023_32}, and ORB \cite{6126544}. Later advancements, including the NetVLAD series \cite{arandjelović2016netvladcnnarchitectureweakly, hausler2021patchnetvladmultiscalefusionlocallyglobal} and AnyLoc \cite{keetha2023anylocuniversalvisualplace}, adopted learning-based models to extract feature maps and combine local features into comprehensive global descriptors.

However, the high performance of most VPR approaches is largely attributed to large-scale training on VPR-specific datasets \cite{keetha2023anylocuniversalvisualplace}. Collecting extensive data for outdoor scenes is relatively straightforward due to natural variations in daylight, weather, and seasons. However, such data collection is more challenging in indoor rooms, making large-scale training on indoor datasets difficult and potentially limiting their effectiveness.
Our approach effectively tackles this challenge by focusing on object-oriented feature representations, allowing us to leverage mature, pre-trained models for object feature learning. This design enables AirRoom to deliver robust performance without requiring any additional training or fine-tuning on specific datasets.
\section{Proposed Approach}
\label{sec:proposed_approach}

We propose a simple yet highly effective pipeline, AirRoom, for room reidentification that leverages multi-level object-oriented information, as shown in \fref{fig:pipeline}. We will now systematically introduce each module of the pipeline, following the sequence of stages in which they are executed.

\subsection{Global Stage}

In this stage, we utilize the Global Feature Extractor to capture global context features, which are derived from the collective presence of objects within the room. These features are then used for Global Retrieval, coarsely selecting semantically similar candidate rooms from the database.

\subsubsection{Global Feature Extractor}
\label{sec:section3.1.1}

Indoor rooms exhibit fewer variations compared to outdoor environments. They lack diverse topographies, such as aerial, subterranean, or underwater features, and do not experience temporal changes like day-night or seasonal variations. Consequently, collecting large datasets for each indoor room is challenging, complicating large-scale training as seen in many VPR methods \cite{arandjelović2016netvladcnnarchitectureweakly, hausler2021patchnetvladmultiscalefusionlocallyglobal, alibey2023mixvprfeaturemixingvisual}. 

However, indoor rooms are inherently rich in objects, each contributing to the room’s overall semantic context. By leveraging this global context information, we can refine the reference search to specifically focus on rooms with similar semantic features to those in the query image. For this purpose, we prefer backbones pretrained on large image datasets, as they provide strong generalizability and effectively capture informative global context features \cite{kornblith2019betterimagenetmodelstransfer}. Our model selections, therefore, include pretrained CNN-based models such as ResNet \cite{he2015deepresiduallearningimage} and transformer-based self-supervised models like DINOv2 \cite{oquab2024dinov2learningrobustvisual}.

\subsubsection{Global Retrieval}

Using the Global Feature Extractor, we extract global context features for \(M\) query and \(N\) reference images. Let \(\mathbf{Q} \in \mathbb{R}^{M \times D_g}\) and \(\mathbf{R} \in \mathbb{R}^{N \times D_g}\) denote the query and reference features, respectively, where \(D_g\) is the feature dimension. The cosine similarity matrix \(\mathbf{S}\) is then computed as:
\begin{equation}
    \mathbf{S}_{ij} = \frac{\mathbf{Q}_i \cdot \mathbf{R}_j}{\|\mathbf{Q}_i\| \|\mathbf{R}_j\|}.
    \label{eq:global feature cosine similarity}
\end{equation}
For each query, we select the top-5 most similar reference candidates using the following formula:
\begin{equation}
    \text{Top}_5(\mathbf{S}_{i, :}) = \text{argsort}(-\mathbf{S}_{i, :})[:5],
    \label{eq:global retrieval}
\end{equation}
where \(\mathbf{S}_{i, :}\) represents the cosine similarity for the \(i\)-th query.

\subsection{Local Stage}

Global context features provide valuable semantic information that helps narrow down the candidate list. However, when faced with many semantically similar rooms, relying solely on global context is insufficient, and local features become increasingly essential. In this stage, we adopt a local perspective by first applying instance segmentation and the Receptive Field Expander to identify objects and patches. We then use the Object Feature Extractor to extract features from both objects and patches, followed by Object-Aware Scoring to further refine the candidate list.

\subsubsection{Instance Segmentation}

For each query image and its corresponding five candidates, we employ instance segmentation methods, such as Mask R-CNN \cite{he2018maskrcnn} and Semantic-SAM \cite{li2023semanticsamsegmentrecognizegranularity}, to identify and delineate individual objects. This process generates each object's mask and bounding box. Next, we calculate the center point \(c\) of each object using its bounding box, as shown below:

\begin{equation}
    c = (\frac{x+W}{2}, \frac{y+H}{2}).
    \label{eq:center point}
\end{equation}
In this equation, \(x\) and \(y\) represent the pixel coordinates of the top-left corner of the bounding box, while \(W\) and \(H\) denote the width and height of the bounding box, respectively.

\subsubsection{Receptive Field Expander}

Single object information alone is not sufficiently discriminative. For example, although different desks may have distinct appearances, they can be found in both dining halls and offices. However, when an object is connected with its neighboring items—such as a desk alongside a computer, keyboard, or notebook—it suggests that the room is more likely to be an office rather than a dining hall. This insight motivates us to expand the receptive field from a single object to a patch containing multiple objects.

Given the center points of all objects in an image, we employ Delaunay triangulation \cite{10.5555/1370949} to generate a triangulated graph of object relationships. Specifically, Delaunay triangulation is applied to the set of object centers, ensuring that no object centers are inside the circumcircle of any triangle. This method maximizes the minimum angle of the triangles, preventing narrow, elongated triangles and ensuring more uniform object adjacency. By analyzing the adjacency relationships among the resulting triangles, we can construct the object adjacency matrix, which encodes the spatial and relational proximity of objects within the room.

\vspace{-10pt}
\begin{figure}[ht]
    \centering
    \includegraphics[width=\columnwidth]{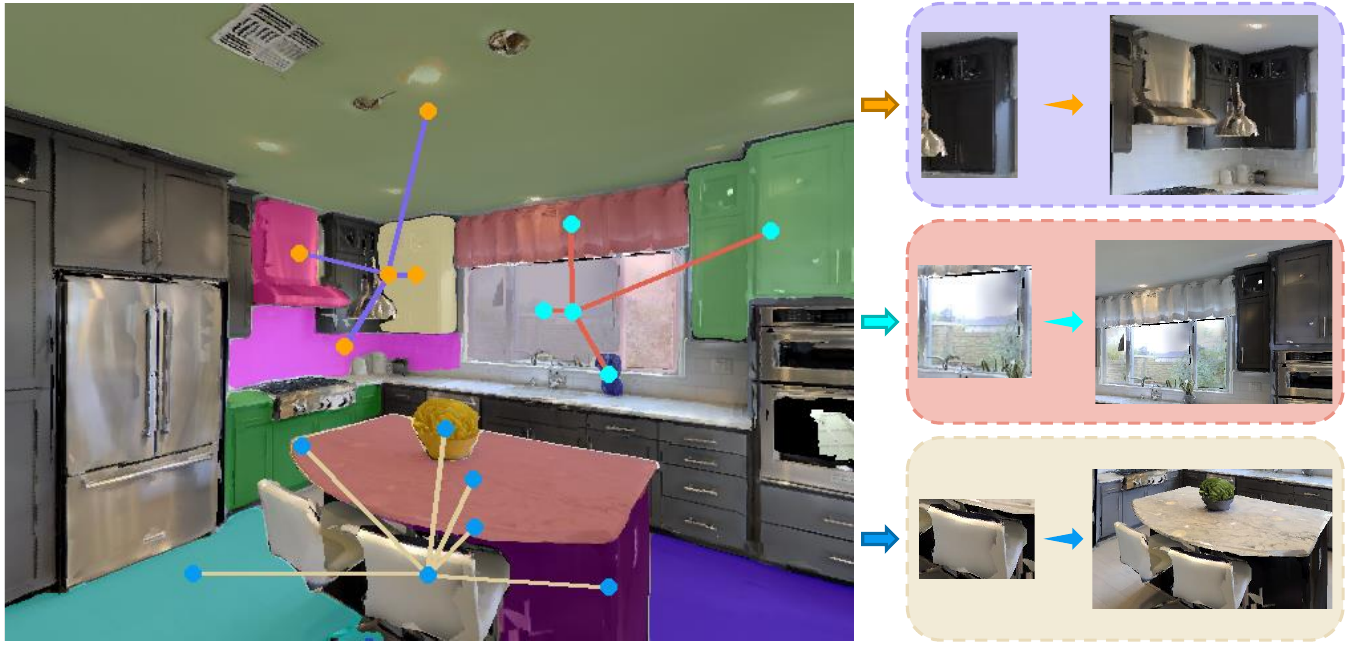}
    \vspace{-20pt}
    \caption{The Receptive Field Expander broadens the receptive field from individual objects to patches rich in contextual information. Leveraging the object adjacency matrix and each object's bounding box, it expands single objects such as a cupboard, window pane, and chair into object patches like a modular kitchen, multi-pane window, and dining set, respectively.}
    \vspace{-5pt}
    \label{fig:expander_image}
\end{figure}


Given the object adjacency matrix and bounding boxes in an image, for each object, we consider the bounding boxes of its neighboring objects and enlarge the current object's bounding box to encompass all adjacent objects. This expansion increases the receptive field, enabling us to capture richer contextual information, as illustrated in \fref{fig:expander_image}. We then apply Non-Maximum Suppression (NMS) to select the highest confidence bounding boxes, removing overlapping ones based on their Intersection over Union (IoU) scores. This results in a set of clean, informative object patches.

\begin{figure*}[t]
    \centering
    \includegraphics[width=\textwidth]{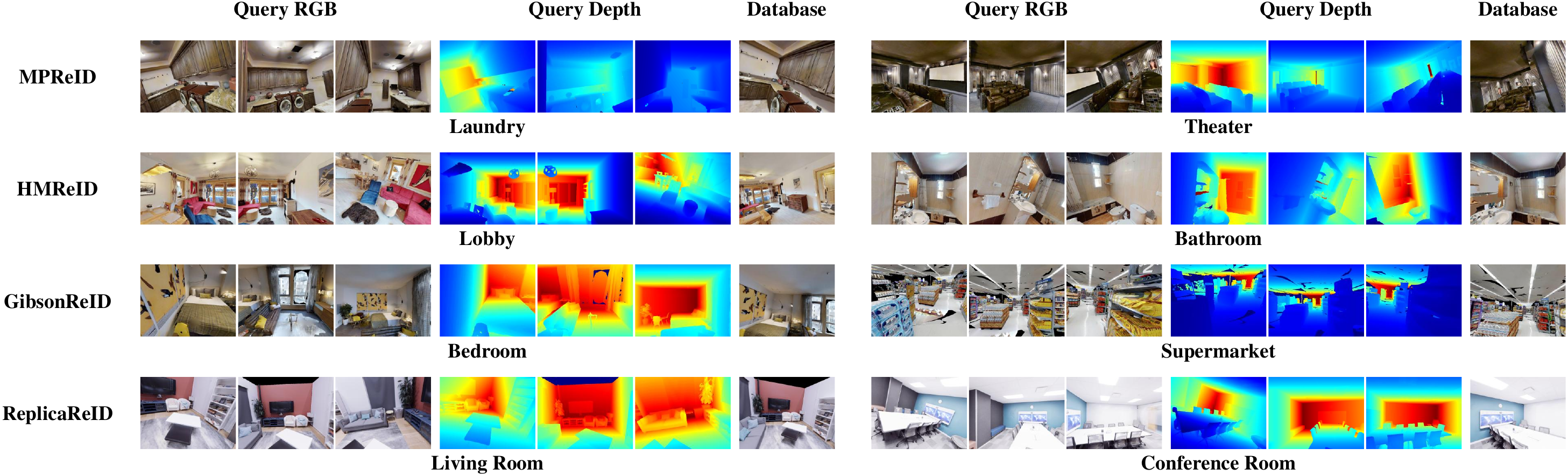}
    \vspace{-20pt}
    \caption{Illustration of four newly constructed room reidentification datasets: MPReID, HMReID, GibsonReID, and ReplicaReID. Each room provides only one reference image in the database, while query images for each room capture varied viewpoints.}
    \vspace{-10pt}
    \label{fig:dataset_image}
\end{figure*}

\subsubsection{Object-Aware Refinement}
\label{subsec:refinement}

The Object-Aware Refinement module is composed of three key submodules: Object Feature Extractor, Mutual Nearest Neighbors, and Object-Aware Scoring.

\vspace{-6pt}
\paragraph{Object Feature Extractor}

To effectively leverage object patches and object segmentation information, we prioritize global features over local feature aggregation. The latter approach may fail to capture object characteristics effectively and can significantly increase computational complexity and storage demands \cite{zheng2018sift}. As discussed in Section~\ref{sec:section3.1.1}, we continue to rely on models pre-trained on large image datasets. Using the Object Feature Extractor, we obtain features for both query and reference patches and objects. Let \(Q_p=\{\mathbf{p_i^q}\}_{i=1}^{n_{qp}}\) and \(Q_o=\{\mathbf{o_i^q}\}_{i=1}^{n_{qo}}\) represent the query patch and object feature sets, respectively. For each reference image among the query’s five \mbox{candidates}, we define the reference patch and object feature sets as \(R_p=\{\mathbf{p_i^r}\}_{i=1}^{n_{rp}}\) and \(R_o=\{\mathbf{o_i^r}\}_{i=1}^{n_{ro}}\).

\vspace{-6pt}
\paragraph{Mutual Nearest Neighbors} Given a set of query features \(\{\mathbf{f_i^q}\}_{i=1}^{n_q}\) and reference features \(\{\mathbf{f_i^r}\}_{i=1}^{n_r}\), we obtain feature pairs by identifying mutual nearest neighbor matches through exhaustive comparison of the two sets. Let \(P\) denote the set of cosine similarity scores for these mutual nearest neighbor matches, then we have
\begin{equation}
    P = \{\cos(\mathbf{f_i^q}, \mathbf{f_j^r}) \mid i = \text{NN}_r(\mathbf{f_j^r}), \; j = \text{NN}_q(\mathbf{f_i^q})\}
    \label{eq:mutual nearest neighbors}
\end{equation}
where
\begin{equation}
    \text{NN}_q(\mathbf{f_i^q}) = \arg\max_{j} \left( \frac{\mathbf{f_i^q} \cdot \mathbf{f_j^r}}{\|\mathbf{f_i^q}\| \|\mathbf{f_j^r}\|} \right),
\end{equation}
\begin{equation}
    \text{NN}_r(\mathbf{f_i^r}) = \arg\max_{j} \left( \frac{\mathbf{f_i^r} \cdot \mathbf{f_j^q}}{\|\mathbf{f_i^r}\| \|\mathbf{f_j^q}\|} \right),
\end{equation}
\begin{equation}
    \cos(\mathbf{f_i^q}, \mathbf{f_j^r}) = \frac{\mathbf{f_i^q} \cdot \mathbf{f_j^r}}{\|\mathbf{f_i^q}\| \|\mathbf{f_j^r}\|}.
\end{equation}
By utilizing mutual nearest neighbors, we can significantly improve retrieval accuracy, simultaneously narrowing the search space and enhancing overall retrieval efficiency \cite{zhong2017reranking}.

\vspace{-6pt}
\paragraph{Object-Aware Scoring} The object-aware score \(s\) is the sum of the global score \(s_{\text{global}}\) (calculated in Equation~\ref{eq:global feature cosine similarity}), the patch score \(s_{\text{patch}}\), and the object score \(s_{\text{object}}\):
\begin{equation}
    s = s_{\text{global}} + s_{\text{patch}}(Q_p, R_p) + s_{\text{object}}(Q_o, R_o).
    \label{eq:object-aware scoring}
\end{equation}
Here, \(s_{\text{patch}}\) and \(s_{\text{object}}\) can either be \(s_{\text{mean}}\) or \(s_{\max}\), where
\begin{subequations}
\begin{align}
    s_{\text{mean}}(Q_t, R_t) &= \frac{1}{|P(Q_t, R_t)|} \sum_{x \in P(Q_t, R_t)} x,
    \label{eq:mean}\\
    s_{\max}(Q_t, R_t) &= \max_{x \in P(Q_t, R_t)} x.
    \label{eq:max}
\end{align}
\end{subequations}
In these equations, \(P\) denotes the set of cosine similarity scores for mutual nearest neighbor matches, with \(Q_t\) representing either \(Q_p\) or \(Q_o\), and \(R_t\) representing either \(R_p\) or \(R_o\). The global score \(s_{\text{global}}\) serves as a prior, indicating that the initial five candidates vary in relevance. Thus, we retain this term to account for their differing levels of relevance.

\vspace{-8pt}
\paragraph{Object-Aware Refinement} For each query, we select the top-2 most similar reference candidates from the initial five using the Object-Aware Scoring:
\begin{equation}
    \text{Top}_2(\mathbf{s}_{i}) = \text{argsort}(-\mathbf{s}_{i})[:2],
\end{equation}
where \(\mathbf{s}_{i}\) is the object-aware scores for the \(i\)-th query.

\subsection{Fine-Grained Stage}

Patch and object features provide valuable information for understanding the room layout; however, they may be insufficient when distinguishing highly visually similar rooms, particularly in the presence of viewpoint variations and occlusions. Keypoints on objects, by contrast, exhibit strong robustness to texture and appearance variations, enabling them to effectively handle partial occlusions and reject outliers \cite{1498756}. This allows keypoints to offer a more refined approach, capturing finer details for more accurate room identification. In this stage, we use Fine-Grained Retrieval to select the final top-1 result.

\subsubsection{Fine-Grained Retrieval}

Deep matchers, such as SuperGlue \cite{sarlin2020supergluelearningfeaturematching}, perform well in visual localization tasks under challenging conditions, both indoors and outdoors. However, they tend to face efficiency issues. In contrast, LightGlue \cite{lindenberger2023lightgluelocalfeaturematching} offers high efficiency without compromising matching accuracy, making it an ideal choice for our Fine-Grained Retrieval.

For each query image and its two candidate reference images, we match the query to each candidate and record the number of matching keypoint pairs. A higher number of matches typically indicates greater overlap and consistency between the features of the two images, suggesting a higher degree of similarity in their content \cite{Lowe2004DistinctiveIF}. The candidate with more matches is selected as the final result.

\vspace{-3pt}
\section{Experimental Results}
\label{sec:experimental_results}

\begin{table*}[t]
\centering
\resizebox{\textwidth}{!}{%
\begin{tabular}{l|cccc|cccc|cccc|cccc}
\toprule
\multirow{2}{*}{\textbf{Methods}} & \multicolumn{4}{c|}{\textbf{MPReID}} & \multicolumn{4}{c|}{\textbf{HMReID}} & \multicolumn{4}{c|}{\textbf{GibsonReID}} & \multicolumn{4}{c}{\textbf{ReplicaReID}} \\
 & Accuracy & Precision & Recall & F1 & Accuracy & Precision & Recall & F1 & Accuracy & Precision & Recall & F1 & Accuracy & Precision & Recall & F1 \\
\midrule
CVNet & 17.45 & 29.52 & 17.45 & 19.34 & 11.71 & 25.42 & 11.95 & 13.86 & 12.04 & 24.06 & 12.07 & 14.27 & 15.93 & 20.53 & 15.74 & 16.64 \\
DINOv2 & 59.36 & 64.68 & 59.36 & 58.91 & 53.91 & 60.52 & 53.73 & 54.69 & 61.01 & 65.88 & 61.78 & 61.71 & 78.06 & 79.68 & 77.97 & 77.44 \\
Patch-NetVLAD & 64.32 & 70.47 & 64.36 & 65.53 & 64.86 & 68.78 & 64.32 & 65.16 & 61.47 & 66.90 & 62.04 & 62.51 & 63.77 & 64.97 & 63.86 & 63.87 \\
AnyLoc & 92.34 & 93.23 & 92.36 & 92.32 & 89.69 & 90.25 & 89.53 & 89.62 & 85.85 & 87.42 & 86.15 & 86.21 & \textbf{88.57} & \textbf{89.89} & \textbf{88.46} & \textbf{88.42} \\
\rowcolor{Lavender}
AirRoom & \textbf{93.96} & \textbf{94.52} & \textbf{93.98} & \textbf{93.91} & \textbf{93.80} & \textbf{94.01} & \textbf{93.55} & \textbf{93.62} & \textbf{91.68} & \textbf{92.41} & \textbf{91.79} & \textbf{91.63} & 87.18 & 89.39 & 87.08 & 87.24 \\
\bottomrule
\end{tabular}%
}
\vspace{-5pt}
\caption{Overall performance comparison between AirRoom and baseline models on four newly constructed room ReID datasets.}
\vspace{-12pt}
\label{tab:overall}
\end{table*}

\subsection{Datasets}
\vspace{-5pt}
No existing indoor scene datasets are ideally suited for room reidentification tasks, as none fully satisfy the requirements.  Datasets like ScanNet++ \cite{yeshwanth2023scannethighfidelitydataset3d} and MIT Indoor Scenes \cite{5206537} lack room-level segmentation, resulting in multiple rooms sharing a single scene label. The 17 Places \cite{7801503} dataset includes uniquely labeled rooms but offers limited viewpoint variations, and the images are often vague. While this dataset also includes day-night changes, these are not particularly relevant for most indoor scenarios. The Reloc110 \cite{aryan2023airlocobjectbasedindoorrelocalization} dataset is likely the most suitable option; however, its quality is insufficient, with many images containing only solid-colored walls or floors due to random sampling, resulting in minimal contextual information.

Several high-quality indoor 3D datasets—such as Matterport3D \cite{Matterport3D}, Habitat-Matterport3D \cite{ramakrishnan2021hm3d}, the Gibson Database of 3D Spaces \cite{xiazamirhe2018gibsonenv}, and Replica \cite{replica19arxiv}—offer real-world indoor scenes. Building on these resources and utilizing the interactive Habitat Simulator \cite{puig2023habitat3, szot2021habitat, habitat19iccv}, we created four new datasets: MPReID, HMReID, GibsonReID, and ReplicaReID, as shown in \fref{fig:dataset_image}.



Using the Habitat Simulator, we configured an agent for each room and manually selected 5 to 10 key poses to guide its exploration. The agent captured 640×480 RGB-D images from various angles, resulting in 300 to 800 images per room, depending on the number of key poses. However, many randomly sampled images were of low quality, often containing only walls or floors with minimal context. To address this, we carefully filtered the images for each room, retaining those that accurately represented the space and provided valuable information for room ReID.

In total, the datasets are as follows: MPReID includes 15 scenes, 105 rooms, and 16,231 RGB-D images; HMReID consists of 21 scenes, 105 rooms, and 15,781 RGB-D images; GibsonReID contains 24 scenes, 45 rooms, and 6,743 RGB-D images; and ReplicaReID includes 12 scenes, 19 rooms, and 2,862 RGB-D images.

\vspace{-4pt}
\subsection{Database Preprocess}
\vspace{-4pt}

In the room reidentification setting, we have multiple query images and a reference database. For each dataset, we select only one image per room to build the database. Specifically, for all the images of each room, we first use CLIP \cite{radford2021learningtransferablevisualmodels} to extract feature embeddings. Then, we apply K-means clustering with the number of clusters set to 1. The image closest to the cluster center is chosen as the reference image, as it best represents the room's visual characteristics \cite{tan2005introduction}.

After building the reference database, we preprocess features. First, we use the Global Feature Extractor to obtain and save the global context features. Next, we apply the instance segmentation module to segment the objects. Then, we use our Receptive Field Expander to obtain object patches and the Object Feature Extractor to extract and save the features of both the objects and the patches.

\vspace{-4pt}
\subsection{Experimental Overview}
\vspace{-4pt}
We conducted five primary experiments: overall performance comparison, group-wise performance comparison, pipeline flexibility evaluation, ablation studies, and runtime analysis. For evaluation, we used accuracy, precision, recall, and the F1 score as metrics. Per-class precision, recall, and F1-score were computed using a multi-class confusion matrix, followed by macro averaging. Accuracy was measured as the ratio of correctly matched queries to the total number of queries. A detailed runtime analysis and additional experimental results are provided in the appendix.


\begin{table*}[t]
\centering
\resizebox{\textwidth}{!}{%
\begin{tabular}{l|cccc|cccc|cccc|cccc}
\toprule
\multirow{2}{*}{\textbf{Methods}} & \multicolumn{4}{c|}{\textbf{MPReID}} & \multicolumn{4}{c|}{\textbf{HMReID}} & \multicolumn{4}{c|}{\textbf{GibsonReID}} & \multicolumn{4}{c}{\textbf{ReplicaReID}} \\
 & Accuracy & Precision & Recall & F1 & Accuracy & Precision & Recall & F1 & Accuracy & Precision & Recall & F1 & Accuracy & Precision & Recall & F1 \\
\midrule
ResNet50 & 76.14 & 79.21 & 76.20 & 76.58 & 69.03 & 73.21 & 68.61 & 69.07 & 68.84 & 72.30 & 69.50 & 69.00 & 75.05 & 78.61 & 75.30 & 74.88 \\
CVNet & 17.45 & 29.52 & 17.45 & 19.34 & 11.71 & 25.42 & 11.95 & 13.86 & 12.04 & 24.06 & 12.07 & 14.27 & 15.93 & 20.53 & 15.74 & 16.64 \\
\rowcolor{Lavender}
AirRoom-ResNet50 & \textbf{86.16} & \textbf{87.69} & \textbf{86.19} & \textbf{86.16} & \textbf{81.23} & \textbf{83.90} & \textbf{80.76} & \textbf{81.23} & \textbf{82.53} & \textbf{84.91} & \textbf{82.86} & \textbf{82.54} & \textbf{83.51} & \textbf{84.85} & \textbf{83.54} & \textbf{83.17} \\
\cdashline{1-17}
NetVLAD & 82.22 & 86.77 & 82.24 & 82.92 & 72.04 & 80.79 & 71.83 & 73.05 & 68.86 & 81.00 & 69.24 & 71.01 & 77.04 & 81.31 & 77.28 & 77.63 \\
Patch-NetVLAD(4096) & 64.32 & 70.47 & 64.36 & 65.53 & 64.86 & 68.78 & 64.32 & 65.16 & 61.47 & 66.90 & 62.04 & 62.51 & 63.77 & 64.97 & 63.86 & 63.87 \\
Patch-NetVLAD(512) & 66.62 & 71.85 & 66.67 & 67.62 & 65.63 & 69.28 & 65.01 & 65.57 & 60.95 & 69.16 & 61.43 & 62.46 & 66.00 & 68.75 & 66.25 & 66.22 \\
Patch-NetVLAD(128) & 65.04 & 70.84 & 65.09 & 66.15 & 61.17 & 66.71 & 60.69 & 61.42 & 58.31 & 66.15 & 58.69 & 59.66 & 61.88 & 66.29 & 62.12 & 62.05 \\
\rowcolor{Lavender}
AirRoom-NetVLAD & \textbf{89.38} & \textbf{90.99} & \textbf{89.40} & \textbf{89.50} & \textbf{83.47} & \textbf{86.91} & \textbf{83.08} & \textbf{83.66} & \textbf{82.29} & \textbf{87.27} & \textbf{82.61} & \textbf{82.98} & \textbf{83.58} & \textbf{84.42} & \textbf{83.60} & \textbf{83.37} \\
\bottomrule
\end{tabular}%
}
\vspace{-6pt}
\caption{Group-wise performance comparison with baseline models to assess the effectiveness of the object-aware mechanism.}
\vspace{-5pt}
\label{tab:grouped}
\end{table*}

\begin{table*}[t]
\centering
\resizebox{\textwidth}{!}{%
\begin{tabular}{l|cccc|cccc|cccc|cccc}
\toprule
\multirow{2}{*}{\textbf{Methods}} & \multicolumn{4}{c|}{\textbf{MPReID}} & \multicolumn{4}{c|}{\textbf{HMReID}} & \multicolumn{4}{c|}{\textbf{GibsonReID}} & \multicolumn{4}{c}{\textbf{ReplicaReID}} \\
 & Accuracy & Precision & Recall & F1 & Accuracy & Precision & Recall & F1 & Accuracy & Precision & Recall & F1 & Accuracy & Precision & Recall & F1 \\
\midrule
ViT & 81.90 & 85.27 & 81.96 & 81.71 & 76.47 & 79.37 & 76.04 & 75.91 & 76.46 & 78.51 & 77.00 & 76.88 & 77.99 & 81.41 & 78.15 & 77.46 \\
\rowcolor{Lavender} AirRoom-ViT & \textbf{89.70} & \textbf{90.97} & \textbf{89.72} & \textbf{89.35} & \textbf{86.58} & \textbf{88.13} & \textbf{86.12} & \textbf{86.23} & \textbf{87.08} & \textbf{88.24} & \textbf{87.33} & \textbf{87.19} & \textbf{84.84} & \textbf{86.85} & \textbf{84.79} & \textbf{84.45} \\
\cdashline{1-17}
DINO & 80.66 & 84.32 & 80.73 & 81.14 & 73.54 & 77.73 & 73.13 & 73.79 & 72.28 & 74.92 & 72.92 & 72.89 & 86.58 & 87.77 & 86.60 & 86.49 \\
\rowcolor{Lavender} AirRoom-DINO & \textbf{88.00} & \textbf{89.59} & \textbf{88.05} & \textbf{88.09} & \textbf{83.62} & \textbf{85.43} & \textbf{83.14} & \textbf{83.40} & \textbf{84.62} & \textbf{86.23} & \textbf{84.95} & \textbf{84.83} & \textbf{87.49} & \textbf{88.56} & \textbf{87.41} & \textbf{87.25} \\
\cdashline{1-17}
DINOv2 & 59.36 & 64.68 & 59.36 & 58.91 & 53.91 & 60.52 & 53.73 & 54.69 & 61.01 & 65.88 & 61.78 & 61.71 & 78.06 & 79.68 & 77.97 & 77.44 \\
\rowcolor{Lavender} AirRoom-DINOv2 & \textbf{76.10} & \textbf{79.03} & \textbf{76.11} & \textbf{75.80} & \textbf{70.95} & \textbf{73.86} & \textbf{70.66} & \textbf{70.78} & \textbf{78.63} & \textbf{80.44} & \textbf{79.00} & \textbf{78.45} & \textbf{85.57} & \textbf{86.58} & \textbf{85.45} & \textbf{85.19} \\
\cdashline{1-17}
AnyLoc(16) & 90.22 & 91.18 & 90.25 & 90.17 & 84.63 & 86.40 & 84.56 & 84.91 & 82.20 & 83.77 & 82.59 & 82.74 & 85.64 & 87.52 & 85.59 & 85.67 \\
\rowcolor{Lavender} AirRoom-AnyLoc(16) & \textbf{93.05} & \textbf{93.66} & \textbf{93.08} & \textbf{92.99} & \textbf{91.55} & \textbf{92.12} & \textbf{91.32} & \textbf{91.47} & \textbf{89.04} & \textbf{89.97} & \textbf{89.21} & \textbf{89.13} & \textbf{86.83} & \textbf{89.03} & \textbf{86.76} & \textbf{86.90} \\
\cdashline{1-17}
AnyLoc(8) & 88.03 & 89.33 & 88.08 & 88.01 & 81.93 & 83.89 & 81.94 & 82.25 & 79.27 & 81.29 & 79.72 & 79.71 & 84.98 & 86.19 & 85.03 & 84.88 \\
\rowcolor{Lavender} AirRoom-AnyLoc(8) & \textbf{92.37} & \textbf{93.14} & \textbf{92.40} & \textbf{92.32} & \textbf{90.24} & \textbf{90.85} & \textbf{90.01} & \textbf{90.13} & \textbf{88.37} & \textbf{89.38} & \textbf{88.56} & \textbf{88.52} & \textbf{85.81} & \textbf{87.67} & \textbf{85.77} & \textbf{85.80} \\
\bottomrule
\end{tabular}%
}
\vspace{-6pt}
\caption{Global Feature Extractor Flexibility.}
\label{tab:global feature extractor flexibility}
\vspace{-15pt}
\end{table*}

\vspace{-4pt}
\subsection{Overall Performance Comparison}
\vspace{-4pt}
\label{sec:section4.4}

In this section, we present a performance comparison between the best-performing version of our approach and several state-of-the-art methods, allowing us to benchmark our pipeline against established room reidentification models across different feature extraction and retrieval strategies.

We selected three categories of baseline methods: image retrieval (CVNet \cite{lee2022correlationverificationimageretrieval}), global descriptor-based visual place recognition (VPR) (DINOv2 \cite{oquab2024dinov2learningrobustvisual}), and VPR using aggregated local features (Patch-NetVLAD \cite{hausler2021patchnetvladmultiscalefusionlocallyglobal} and AnyLoc \cite{keetha2023anylocuniversalvisualplace}). Specifically, we used the Base version of DINOv2, configured CVNet with a ResNet50 \cite{he2015deepresiduallearningimage} backbone and a reduction dimension of 2048, selected the performance version of Patch-NetVLAD, and set up AnyLoc with AnyLoc-VLAD-DINOv2 using 32 VLAD clusters.


\begin{figure}[ht]
    \centering
    \includegraphics[width=\columnwidth]{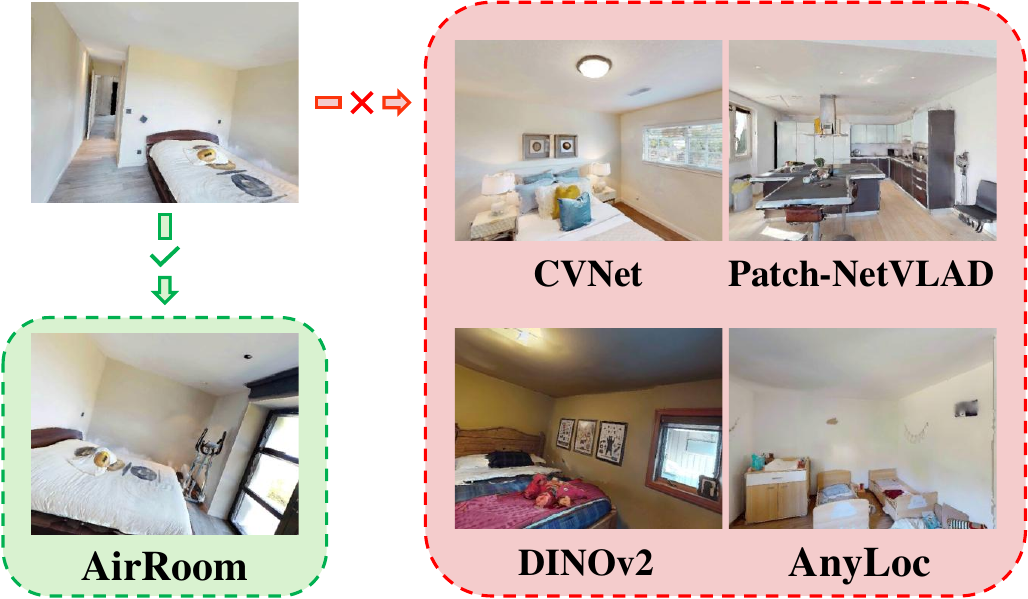}
    \vspace{-16pt}
    \caption{Given a bedroom query, AirRoom accurately retrieves the target image by leveraging object relevance for room reidentification. In contrast, CVNet retrieves visually similar images without preserving scene accuracy, DINOv2 captures semantic content but overlooks color details, Patch-NetVLAD, using aggregated local features to form global descriptors, retrieves images with mismatched semantic information, and AnyLoc considers semantic and color attributes but neglects object importance within rooms.}
    \vspace{-5pt}
    \label{fig:failure}
\end{figure}

\tref{tab:overall} presents a quantitative comparison between AirRoom and baseline methods, showing that AirRoom outperforms all baselines on nearly all metrics and datasets. In room reidentification tasks, image retrieval methods generally exhibit lower classification metrics due to their focus not being on top-1 precision, while VPR methods yield better results. Global descriptor-based VPR methods capture only high-level semantic information, often retrieving rooms with similar semantics but lacking detailed features. In contrast, VPR methods using aggregated local features, such as Patch-NetVLAD, emphasize low-level encodings but may overlook global context, resulting in less accurate retrievals. \fref{fig:failure} illustrates failure cases for CVNet, DINOv2, Patch-NetVLAD, and AnyLoc, highlighting these limitations. 
Although AnyLoc, known for its robust performance in ``anywhere, anytime, anyview" VPR, performs well, AirRoom further enhances performance, achieving a 20\% to 40\% improvement within the available margin compared to AnyLoc. For instance, AnyLoc achieves 89.69\% accuracy on HMReID, leaving approximately 10\% room for improvement. AirRoom, with an accuracy of 93.80\%, demonstrates up to a 40\% improvement within this remaining margin. These results highlight AirRoom's superior precision and refinement in room reidentification.

\subsection{Group-Wise Performance Comparison}
\label{sec:section4.5}
\vspace{-3pt}
Many baseline methods adopt a “backbone + enhancement mechanism” paradigm, which our approach also follows. In this section, we compare the performance of our object-aware enhancement mechanism with that of several state-of-the-art methods, using the same backbone as each group’s baseline. This setup allows us to directly assess the effectiveness of our object-aware enhancement mechanism.

For the ResNet50 backbone group, we use CVNet \cite{lee2022correlationverificationimageretrieval} as the baseline. In the NetVLAD backbone group, we employ Patch-NetVLAD \cite{hausler2021patchnetvladmultiscalefusionlocallyglobal} as the baseline, testing it at three reduction dimensions: 4096, 512, and 128.

\tref{tab:grouped} reveals that within each group, the single backbone outperforms the baseline methods that attempt to enhance performance through various mechanisms, indicating that these mechanisms do not effectively capture critical information in indoor rooms. In contrast, our object-aware enhancement mechanism significantly improves the backbone’s performance by emphasizing the importance of \mbox{objects} in indoor environments.

\subsection{Pipeline Flexibility Evaluation}
\label{sec:section4.6}
\vspace{-3pt}
In this section, we systematically evaluate the flexibility and adaptability of AirRoom by testing different configurations of its key modules. The results clearly demonstrate that AirRoom is not reliant on any specific model and can effectively integrate a diverse range of models.

\vspace{-5pt}
\subsubsection{Global Feature Extractor}
\vspace{-4pt}
We test various Global Feature Extractors, including ViT \cite{dosovitskiy2021imageworth16x16words}, DINO \cite{caron2021emergingpropertiesselfsupervisedvision}, DINOv2 \cite{oquab2024dinov2learningrobustvisual}, and AnyLoc-VLAD-DINOv2 \cite{keetha2023anylocuniversalvisualplace} with VLAD cluster sizes of 16 and 8.

As shown in \tref{tab:global feature extractor flexibility}, AirRoom consistently achieves over 85\% across all metrics and datasets in nearly every case, regardless of the capabilities of the Global Feature Extractor used. Even in the single exception with DINOv2, AirRoom still improves performance by nearly 15\%. This demonstrates that the effectiveness of our pipeline is not reliant on any specific Global Feature Extractor, highlighting AirRoom's adaptability to various backbone configurations and underscoring its robust flexibility.

\vspace{-5pt}
\subsubsection{Instance Segmentation}
\vspace{-4pt}
We compare traditional instance segmentation methods, such as Mask R-CNN \cite{he2018maskrcnn}, with more recent approaches, including Semantic-SAM \cite{li2023semanticsamsegmentrecognizegranularity}, which leverage advanced techniques for more granular segmentation.

\tref{tab:is flexibility} shows that AirRoom consistently outperforms the baseline by over 15\%, regardless of the instance segmentation module used. This demonstrates that our pipeline is not dependent on any specific instance segmentation method, underscoring its adaptability in this component.

\begin{table}[h]
\vspace{-5pt}
\centering
\resizebox{\columnwidth}{!}{%
\begin{tabular}{l|cccc}
\toprule
\multirow{2}{*}{\textbf{Methods}} & \multicolumn{4}{c}{\textbf{HMReID}} \\
 & Accuracy & Precision & Recall & F1 \\
\midrule
DINOv2 & 53.91 & 60.52 & 53.73 & 54.69 \\
\rowcolor{Lavender} AirRoom-MaskRCNN & 69.44 & 72.23 & 69.08 & 69.07 \\
\rowcolor{Lavender} AirRoom-SSAM & \textbf{70.95} & \textbf{73.86} & \textbf{70.66} & \textbf{70.78} \\
\bottomrule
\end{tabular}%
}
\vspace{-10pt}
\caption{Instance Segmentation Flexibility.}
\vspace{-16pt}
\label{tab:is flexibility}
\end{table}

\subsubsection{Object Feature Extractor}

We experiment with both traditional backbones, such as ResNet50 \cite{he2015deepresiduallearningimage}, and more modern backbones, like DINOv2 \cite{oquab2024dinov2learningrobustvisual}, as the Object Feature Extractor.

As shown in \tref{tab:ofe flexibility}, AirRoom achieves substantial performance improvements over the baseline, with minimal performance variation between different Object Feature Extractors. This supports the flexibility of our pipeline in accommodating a range of feature extraction methods.

\begin{table}[h]
\vspace{-6pt}
\centering
\resizebox{\columnwidth}{!}{%
\begin{tabular}{l|cccc}
\toprule
\multirow{2}{*}{\textbf{Methods}} & \multicolumn{4}{c}{\textbf{HMReID}} \\
 & Accuracy & Precision & Recall & F1 \\
\midrule
DINOv2 & 53.91 & 60.52 & 53.73 & 54.69 \\
\rowcolor{Lavender} AirRoom-ResNet50 & \textbf{70.95} & \textbf{73.86} & \textbf{70.66} & \textbf{70.78} \\
\rowcolor{Lavender} AirRoom-DINOv2 & 68.67 & 71.81 & 68.33 & 68.59 \\
\bottomrule
\end{tabular}%
}
\vspace{-10pt}
\caption{Object Feature Extractor Flexibility.}
\vspace{-20pt}
\label{tab:ofe flexibility}
\end{table}

\subsubsection{Object-Aware Scoring}

We evaluate both the mean (\(s_{\text{mean}}\)) and max (\(s_{\max}\)) strategies for computing the patch score (\(s_{\text{patch}}\)) and object score (\(s_{\text{object}}\)), assessing their impact on the overall performance.

\tref{tab:os flexibility} shows that AirRoom’s performance remains stable regardless of the object-aware scoring method used. This underscores the robustness of object-oriented information in room reidentification and demonstrates AirRoom’s flexibility in adapting to different scoring strategies.

\begin{table}[h]
\vspace{-6pt}
\centering
\resizebox{\columnwidth}{!}{%
\begin{tabular}{l|cccc}
\toprule
\multirow{2}{*}{\textbf{Methods}} & \multicolumn{4}{c}{\textbf{HMReID}} \\
 & Accuracy & Precision & Recall & F1 \\
\midrule
DINOv2 & 53.91 & 60.52 & 53.73 & 54.69 \\
\rowcolor{Lavender} AirRoom-Max(patch)-Mean(object) & 70.95 & 73.86 & 70.66 & 70.78 \\
\rowcolor{Lavender} AirRoom-Max(patch)-Max(object) & \textbf{71.02} & \textbf{74.02} & \textbf{70.72} & \textbf{70.85} \\
\rowcolor{Lavender} AirRoom-Mean(patch)-Max(object) & 70.85 & 73.85 & 70.55 & 70.70 \\
\rowcolor{Lavender} AirRoom-Mean(patch)-Mean(object) & 70.90 & 73.78 & 70.62 & 70.73 \\
\bottomrule
\end{tabular}%
}
\vspace{-10pt}
\caption{Object-Aware Scoring Flexibility.}
\vspace{-20pt}
\label{tab:os flexibility}
\end{table}

\subsection{Ablation Studies}
\label{sec:section4.7}

In this section, we remove certain modules from our pipeline—including the global score \(s_{\text{global}}\),  the patch score \(s_{\text{patch}}\), the object score \(s_{\text{object}}\), within object-aware scoring, and the entire Fine-Grained Retrieval (FGR)—to assess the importance and effectiveness of each component.

\begin{table}[t]
\centering
\resizebox{\columnwidth}{!}{%
\begin{tabular}{l|cccc}
\toprule
\multirow{2}{*}{\textbf{Methods}} & \multicolumn{4}{c}{\textbf{HMReID}} \\
 & Accuracy & Precision & Recall & F1 \\
\midrule
DINOv2 (AirRoom-w/o all)& 53.91 & 60.52 & 53.73 & 54.69 \\
\rowcolor{Lavender} AirRoom-w/o \(s_{\text{patch}}\) & 66.68 & 70.04 & 66.42 & 66.68 \\
\rowcolor{Lavender} AirRoom-w/o \(s_{\text{object}}\) & 69.77 & 72.84 & 69.48 & 69.64 \\
\rowcolor{Lavender} AirRoom-w/o FGR & 66.11 & 70.85 & 65.80 & 66.41 \\
\rowcolor{Lavender} AirRoom-w/o \(s_{\text{patch}}\) \& \(s_{\text{object}}\) & 62.26 & 66.43 & 62.03 & 62.46 \\
\rowcolor{Lavender} AirRoom-w/o \(s_{\text{patch}}\) \& FGR & 59.39 & 65.25 & 59.14 & 59.97 \\
\rowcolor{Lavender} AirRoom-w/o \(s_{\text{object}}\) \& FGR & 63.44 & 68.68 & 63.14 & 63.84 \\
\rowcolor{Lavender} AirRoom & \textbf{70.95} & \textbf{73.86} & \textbf{70.66} & \textbf{70.78} \\
\bottomrule
\end{tabular}%
}
\vspace{-10pt}
\caption{Ablation Studies (Excluding Global Score Experiments).}
\vspace{-6pt}
\label{tab:ablation w/o global score}
\end{table}

\begin{table}[ht]
\centering
\resizebox{\columnwidth}{!}{%
\begin{tabular}{l|cccc}
\toprule
\multirow{2}{*}{\textbf{Methods}} & \multicolumn{4}{c}{\textbf{HMReID}} \\
 & Accuracy & Precision & Recall & F1 \\
\midrule
ViT & 76.47 & 79.37 & 76.04 & 75.91 \\
\rowcolor{Lavender} AirRoom-ViT-w/o \(s_{\text{global}}\) & 84.86 & 86.82 & 84.34 & 84.61 \\
\rowcolor{Lavender} AirRoom-ViT & \textbf{86.58} & \textbf{88.13} & \textbf{86.12} & \textbf{86.23} \\
\cdashline{1-5}
DINOv2 & 53.91 & 60.52 & 53.73 & 54.69 \\
\rowcolor{Lavender} AirRoom-DINOv2-w/o \(s_{\text{global}}\) & \textbf{71.73} & \textbf{74.97} & \textbf{71.44} & \textbf{71.64} \\
\rowcolor{Lavender} AirRoom-DINOv2 & 70.95 & 73.86 & 70.66 & 70.78 \\
\bottomrule
\end{tabular}%
}
\vspace{-10pt}
\caption{Ablation Studies on Global Score.}
\vspace{-15pt}
\label{tab:ablation on global score}
\end{table}

\tref{tab:ablation w/o global score} shows that removing any module from our pipeline leads to a performance drop. However, as long as at least one module remains, our pipeline still outperforms the baseline. \tref{tab:ablation on global score}  demonstrates that when the Global Feature Extractor (ViT) performs well, the global score \(s_{\text{global}}\) significantly enhances performance. On the other hand, when the Global Feature Extractor (DINOv2) is less effective, the global score \(s_{\text{global}}\) has a slight negative impact, causing a small drop in performance. This result aligns with our hypothesis in Section~\ref{subsec:refinement}, where the global score acts as a prior to rank the priority of the five candidates. Overall, these ablation studies confirm that every module in our pipeline is both important and necessary.

\subsection{Limitations}

While AirRoom achieves state-of-the-art performance in room reidentification under various viewpoint variations, a limitation of our work is the inability to verify robustness to indoor object rearrangements caused by movable objects. Although our mutual nearest neighbors-based Object-Aware Scoring method is somewhat robust to such rearrangements, the datasets used in our experiments lack these cases. In contrast, recent advances in dynamic scene understanding \cite{zhao2024dynamicsceneunderstandingobjectcentric} focus on recognizing scenes in the presence of moving objects, potentially offering greater robustness than our approach. Future work should consider constructing datasets that include object rearrangements and integrating new techniques to enhance robustness to movable objects, thereby improving room reidentification.
\vspace{-5pt}
\section{Conclusion}
\vspace{-5pt}
\label{sec:conclusion}

Room reidentification is a challenging yet crucial research area, with growing applications in fields like augmented reality and homecare robotics. In this paper, we introduce AirRoom, a training-free, object-aware approach for room reidentification. AirRoom leverages multi-level object-oriented features to capture both spatial and contextual information of indoor rooms. To evaluate AirRoom, we constructed four novel datasets specifically for room reidentification. Experimental results demonstrate its robustness to viewpoint variations and superior performance over state-of-the-art methods across nearly all metrics and datasets. Furthermore, the pipeline is highly flexible, maintaining high performance without relying on specific model configurations. Collectively, our work establishes AirRoom as a powerful and versatile solution for precise room reidentification, with broad potential for real-world applications.

\begin{center}
\textbf{Acknowledgments}  
\end{center}
\begin{sloppypar}
\noindent This work was supported by the DARPA award HR00112490426. Any opinions, findings, conclusions, or recommendations expressed in this paper are those of the authors and do not necessarily reflect the views of DARPA.
\end{sloppypar}

{
    \small
    \bibliographystyle{ieeenat_fullname}
    \bibliography{main}
}

\clearpage
\setcounter{page}{1}
\maketitlesupplementary

\section{Datasets}
Table \ref{tab:MPReID} presents the composition of MPReID, while Table \ref{tab:HMReID}, Table \ref{tab:GibsonReID}, and Table \ref{tab:ReplicaReID} outline the compositions of HMReID, GibsonReID, and ReplicaReID, respectively. Table \ref{tab:Statistics} reports the number of semantically different rooms in each room ReID dataset.

\begin{table}[ht]
\centering
\vspace{-3pt}
\resizebox{\columnwidth}{!}{
\begin{tabular}{c|c|c|c|c|c}
\toprule
\textbf{Scene} & \textbf{Rooms} & \textbf{Images} & \textbf{Scene} & \textbf{Rooms} & \textbf{Images} \\
\midrule
8WUmhLawc2A & 8 & 1232 & EDJbREhghzL & 7 & 1078 \\
RPmz2sHmrrY & 5 & 770 & S9hNv5qa7GM & 9 & 1423 \\
ULsKaCPVFJR & 5 & 780 & VzqfbhrpDEA & 7 & 1078 \\
WYY7iVyf5p8 & 4 & 616 & X7HyMhZNoso & 7 & 1078 \\
YFuZgdQ5vWj & 7 & 1078 & i5noydFURQK & 7 & 1078 \\
jh4fc5c5qoQ & 5 & 770 & mJXqzFtmKg4 & 9 & 1386 \\
qoiz87JEwZ2 & 8 & 1232 & wc2JMjhGNzB & 11 & 1708 \\
yqstnuAEVhm & 6 & 924 & \textbf{Total} & \textbf{105} & \textbf{16231} \\
\bottomrule
\end{tabular}
}
\vspace{-1em}
\caption{Composition of MPReID.}
\label{tab:MPReID}
\end{table}

\vspace{-1em}

\begin{table}[ht]
\centering
\resizebox{\columnwidth}{!}{
\begin{tabular}{c|c|c|c|c|c}
\toprule
\textbf{Scene} & \textbf{Rooms} & \textbf{Images} & \textbf{Scene} & \textbf{Rooms} & \textbf{Images} \\
\midrule
7dmR22gwQpH & 6 & 924 & ACZZiU6BXLz & 5 & 682 \\
CETmJJqkhcK & 5 & 813 & CFVBbU9Rsyb & 5 & 770 \\
Coer9RdivP7 & 3 & 462 & DZsJKHoqEYg & 5 & 793 \\
EQSguCqe5Rk & 5 & 819 & Fgtk7tL8R9Y & 5 & 822 \\
GLAQ4DNUx5U & 7 & 1156 & GcfUJ79xCZc & 5 & 572 \\
NcK5aACg44h & 5 & 754 & P8L1328HrLi & 5 & 819 \\
VSxVP19Cdyw & 5 & 769 & b3CuYvwpzZv & 5 & 690 \\
ixTj1aTMup2 & 5 & 757 & ochRmQAHtkF & 5 & 641 \\
qWb4MVxqCW7 & 6 & 879 & rrjjmoZhZCo & 5 & 704 \\
w7QyjJ3H9Bp & 5 & 692 & zR6kPe1PsyS & 5 & 803 \\
zepmXAdrpjR & 3 & 460 & \textbf{Total} & \textbf{105} & \textbf{15781} \\
\bottomrule
\end{tabular}
}
\vspace{-1em}
\caption{Composition of HMReID.}
\label{tab:HMReID}
\end{table}

\vspace{-1em}

\begin{table}[ht]
\centering
\resizebox{\columnwidth}{!}{
\begin{tabular}{c|c|c|c|c|c}
\toprule
\textbf{Scene} & \textbf{Rooms} & \textbf{Images} & \textbf{Scene} & \textbf{Rooms} & \textbf{Images} \\
\midrule
Ackermanville & 1 & 154 & Angiola & 1 & 154 \\
Avonia & 2 & 308 & Beach & 3 & 462 \\
Branford & 1 & 154 & Brevort & 1 & 154 \\
Cason & 2 & 262 & Cooperstown & 2 & 308 \\
Corder & 2 & 308 & Creede & 4 & 526 \\
Elmira & 2 & 308 & Eudora & 2 & 308 \\
Fredericksburg & 2 & 308 & Greigsville & 1 & 154 \\
Idanha & 1 & 154 & Laytonsville & 3 & 462 \\
Lynxville & 2 & 308 & Mahtomedi & 2 & 257 \\
Mayesville & 2 & 308 & Northgate & 1 & 154 \\
Ogilvie & 2 & 308 & Ophir & 3 & 462 \\
Pablo & 1 & 154 & Sumas & 2 & 308 \\
- & - & - & \textbf{Total} & \textbf{45} & \textbf{6743} \\
\bottomrule
\end{tabular}
}
\vspace{-1em}
\caption{Composition of GibsonReID.}
\label{tab:GibsonReID}
\end{table}

\vspace{-1em}

\begin{table}[ht]
\centering
\resizebox{\columnwidth}{!}{
\begin{tabular}{c|c|c|c|c|c}
\toprule
\textbf{Scene} & \textbf{Rooms} & \textbf{Images} & \textbf{Scene} & \textbf{Rooms} & \textbf{Images} \\
\midrule
apartment\_0 & 3 & 462 & apartment\_1 & 1 & 154 \\
apartment\_2 & 4 & 616 & frl\_apartment\_0 & 3 & 426 \\
hotel\_0 & 1 & 154 & office\_0 & 1 & 154 \\
office\_2 & 1 & 140 & office\_3 & 1 & 140 \\
office\_4 & 1 & 154 & room\_0 & 1 & 154 \\
room\_1 & 1 & 154 & room\_2 & 1 & 154 \\
- & - & - & \textbf{Total} & \textbf{19} & \textbf{2862} \\
\bottomrule
\end{tabular}
}
\vspace{-1em}
\caption{Composition of ReplicaReID.}
\label{tab:ReplicaReID}
\end{table}

\begin{table}[h]
\centering
\resizebox{\columnwidth}{!}{
\begin{tabular}{c|c|c|c|c|c|c|c|c|c|c|c|c|c|c}
\hline
 & bathroom & kitchen & living & office & bedroom & theater & dining & wardrobe & gym & laundry & garage & storage & nursery & supermarket \\
\hline
MPReID & 13 & 13 & 20 & 3 & 41 & 4 & 4 & 2 & 2 & 2 & 1 & 0 & 0 & 0 \\
\hline
HMReID & 10 & 18 & 29 & 8 & 31 & 0 & 3 & 1 & 0 & 1 & 0 & 2 & 2 & 0 \\
\hline
GibsonReID & 2 & 10 & 11 & 3 & 12 & 0 & 1 & 0 & 3 & 1 & 0 & 1 & 0 & 1 \\
\hline
ReplicaReID & 0 & 2 & 6 & 6 & 3 & 0 & 2 & 0 & 0 & 0 & 0 & 0 & 0 & 0 \\
\hline
\end{tabular}
}
\caption{Statistics of semantically different rooms across four newly constructed room ReID datasets.}
\label{tab:Statistics}
\end{table}

\section{Experimental Details}

\subsection{Overall Performance Comparison}
\label{sec:appendix_overall}

\paragraph{Baseline Configuration} For CVNet, we use ResNet50 as the backbone and set the reduction dimension to 2048. For DINOv2, we utilize the DINOv2-Base checkpoint. For Patch-NetVLAD, we load pre-trained weights optimized on the Pittsburgh dataset, apply WPCA to reduce feature embedding dimensionality to 4096, set RANSAC as the matcher, use patch weights of 0.45, 0.15, and 0.4, configure patch sizes to 2, 5, and 8 with strides of 1 for all. For AnyLoc, we adopt AnyLoc-VLAD-DINOv2 with the DINOv2 ViT-G/14 architecture, set the descriptor layer to 31, use VLAD with 32 clusters, and specify the domain as \mbox{indoor}.

\paragraph{Baseline Adaptation} For CVNet and Patch-NetVLAD, we perform global retrieval by selecting the top-5 candidates, followed by re-ranking. For CVNet, the candidate with the highest CVNet-Rerank image similarity score is chosen as the final result, while for Patch-NetVLAD, the reference with the highest RANSAC score in the Pairwise Local Matching stage is selected. For DINOv2 and AnyLoc, global features are extracted from the query and reference images, and cosine similarity is computed. The reference image with the highest cosine similarity score is selected as the final match.

\paragraph{AirRoom Configuration} For the Global Feature Extractor, we use AnyLoc-VLAD-DINOv2 with the DINOv2 ViT-G/14 architecture, setting the descriptor layer to 31, applying VLAD with 32 clusters, and specifying the domain as indoor. For Instance Segmentation, we employ Semantic-SAM with pre-trained weights from SA-1B and a SwinL backbone. The Object Feature Extractor is implemented using a ResNet50 model pre-trained on the ImageNet dataset. For Fine-Grained Retrieval, we use LightGlue with the maximum number of keypoints set to 2048.

\subsection{Group-Wise Performance Comparison}

\paragraph{Baseline Configuration} For the ResNet50 backbone group, the configurations for ResNet50 and CVNet follow those detailed in \sref{sec:appendix_overall}. For the NetVLAD backbone group, we use NetVLAD with VGG-16 as the feature extractor, configured with 64 clusters and a feature dimensionality of 512. For Patch-NetVLAD, the feature dimensionalities are set to 4096, 512, and 128, respectively, with all other settings consistent with \sref{sec:appendix_overall}.

\paragraph{Baseline Adaptation} For the ResNet50 backbone group, ResNet50 extracts global features from the query and reference images, with cosine similarity used to select the reference image with the highest score as the final match. The adaptation for CVNet is detailed in \sref{sec:appendix_overall}. For the NetVLAD backbone group, NetVLAD aggregates global descriptors from the query and reference local features, and the reference with the highest cosine similarity score is chosen as the final result. The adaptation for Patch-NetVLAD also follows \sref{sec:appendix_overall}.

\paragraph{AirRoom Configuration} For the ResNet50 backbone group, ResNet50 is used as the Global Feature Extractor, with the configuration consistent with \sref{sec:appendix_overall}. For the NetVLAD backbone group, NetVLAD is used as the Global Feature Extractor, following the configuration outlined in the Baseline Configuration paragraph in this section. The configurations for the remaining modules in both groups are also consistent with \sref{sec:appendix_overall}.

\subsection{Pipeline Flexibility Evaluation}

\subsubsection{Global Feature Extractor}

\paragraph{Baseline Configuration} For ViT, we use the Base variant with a patch size of 16 and an input image size of 224×224, loading pre-trained weights from ImageNet. For DINO, we adopt the DINO-pretrained Vision Transformer Small (ViT-S/16) variant. The configuration for DINOv2 follows \sref{sec:appendix_overall}. For AnyLoc, VLAD clusters are set to 16 and 8, with all other configurations consistent with \sref{sec:appendix_overall}.

\paragraph{Baseline Adaptation} All baselines are used to extract features from query and reference images, with cosine similarity computed to identify the reference room with the highest similarity score.

\paragraph{AirRoom Configuration} For comparisons with a backbone baseline, the backbone is used as the Global Feature Extractor. Backbone configurations follow those outlined in the Baseline Configuration paragraph of this section, while the configurations for the remaining modules in AirRoom are consistent with \sref{sec:appendix_overall}.

\subsubsection{Instance Segmentation}

\paragraph{AirRoom Configuration} DINOv2 is used as the Global Feature Extractor. For Mask R-CNN, we use Mask R-CNN with a ResNet50 backbone and FPN, loading pre-trained weights trained on COCO. For Semantic-SAM, we employ Semantic-SAM with pre-trained weights from SA-1B and a SwinL backbone. The configurations for the remaining modules are consistent with \sref{sec:appendix_overall}.

\section{Large-Scale Evaluation}
Since the four room ReID datasets were curated in a consistent format, we evaluate our method on their union, resulting in more examples for each room type and assessing the feasibility of the proposed method when scaling the data. To this end, we construct a large-scale dataset, UnionReID, by combining all four datasets. Table \ref{tab:Union} presents a performance comparison between AirRoom and four baseline methods, demonstrating that AirRoom continues to outperform them under large-scale conditions.

\begin{table}[h]
\centering
\resizebox{\columnwidth}{!}{
\begin{tabular}{l|cccc}
\toprule
\multirow{2}{*}{\textbf{Methods}} & \multicolumn{4}{c}{\textbf{UnionReID}} \\
 & Accuracy & Precision & Recall & F1 \\
\midrule
CVNet & 14.10 & 27.53 & 14.10 & 16.19 \\
DINOv2 & 53.01 & 59.44 & 53.02 & 53.50 \\
Patch-NetVLAD & 61.15 & 67.53 & 61.04 & 62.31 \\
AnyLoc & 88.28 & 89.62 & 88.22 & 88.32 \\
\rowcolor{Lavender} AirRoom & \textbf{91.87} & \textbf{92.55} & \textbf{91.76} & \textbf{91.76} \\
\bottomrule
\end{tabular}
}
\caption{Comparison with baseline models on UnionReID to evaluate AirRoom's performance under data scaling.}
\label{tab:Union}
\end{table}

\section{Evaluation on Indoor Localization Datasets}
Strictly speaking, room ReID is a novel task with no previously established datasets and is fundamentally distinct from indoor localization. To address this gap, we introduced four new datasets. However, after reviewing existing indoor localization datasets, we identified two that are marginally usable: InLoc  \cite{taira2018inlocindoorvisuallocalization} and Structured3D \cite{Structured3D}. InLoc \cite{taira2018inlocindoorvisuallocalization} employs area-based rather than room-based splits, with some images capturing only corridors and corners. Structured3D \cite{Structured3D} contains tens of thousands of room instances, but each room has fewer than six viewpoints. These limitations reduce the suitability of these two datasets, though they remain partially usable. Nonetheless, evaluating our method on them can further reinforce its validation.

Table \ref{tab:Indoor} presents the comparison results on the two indoor localization datasets, where AirRoom continues to outperform other methods. Additionally, as InLoc represents a more realistic real-world setting, the results further demonstrate AirRoom's robustness in practical environments.

\begin{table}[h]
\vspace{-7pt}
\centering
\resizebox{\columnwidth}{!}{
\begin{tabular}{l|cccc|cccc}
\toprule
\multirow{2}{*}{\textbf{Methods}} & \multicolumn{4}{c|}{\textbf{InLoc}} & \multicolumn{4}{c}{\textbf{Structured3D}} \\
 & Acc & Prec & Rec & F1 & Acc & Prec & Rec & F1 \\
\midrule
CVNet & 8.41 & 12.49 & 8.41 & 8.99 & 12.60 & 21.39 & 12.60 & 14.22 \\
DINOv2 & 11.13 & 19.93 & 11.13 & 11.85 & 53.00 & 63.60 & 53.00 & 54.04 \\
Patch-NetVLAD & 12.78 & 19.59 & 12.78 & 13.73 & 56.30 & 67.67 & 56.30 & 57.71 \\
AnyLoc & 15.78 & 26.11 & 15.78 & 17.04 & 73.40 & 79.75 & 73.40 & 73.90 \\
\rowcolor{Lavender} AirRoom & \textbf{16.80} & \textbf{26.36} & \textbf{16.80} & \textbf{18.05} & \textbf{76.20} & \textbf{82.88} & \textbf{76.20} & \textbf{76.70} \\
\bottomrule
\end{tabular}
}
\caption{Comparison with baseline models on existing datasets to further validate our method.}
\label{tab:Indoor}
\end{table}

\section{Runtime Analysis}

In this section, we evaluate the runtime of each module and compare the total runtime of our pipeline with several state-of-the-art methods to assess the efficiency of our approach.

\begin{table}[ht]
\centering
\resizebox{\columnwidth}{!}{%
\begin{tabular}{l|cccccc}
\toprule
\multirow{2}{*}{\textbf{Modules}} & \multicolumn{6}{c}{\textbf{Runtime (ms)}} \\
 & t=0 & t=0.1 & t=0.2 & t=0.3 & t=0.4 & t=0.5 \\
\midrule
Global Feature Extractor & 48.8 & 44.1 & 43.2 & 44.0 & 43.0 & 43.8 \\
Global Retrieval & 0.1 & 0.1 & 0.1 & 0.1 & 0.1 & 0.1 \\
Instance Segmentation & 38.7 & 38.1 & 38.2 & 38.1 & 38.0 & 38.0 \\
Receptive Field Expander & 6.9 & 2.9 & 1.7 & 1.3 & 0.9 & 0.7 \\
Object Feature Extractor & 113.7 & 71.3 & 47.0 & 33.3 & 29.0 & 22.8 \\
Object-Aware Scoring & 2.9 & 2.2 & 1.7 & 1.5 & 1.4 & 1.2 \\
Fine-Grained Retrieval & 87.4 & 86.3 & 86.1 & 86.1 & 85.8 & 86.2 \\
\midrule
\textbf{Total} & \textbf{299.9} & \textbf{246.5} & \textbf{219.4} & \textbf{205.7} & \textbf{199.5} & \textbf{194.2} \\
\bottomrule
\end{tabular}%
}
\caption{Mask R-CNN \& ResNet Runtime.}
\label{tab:module_runtime_mr}
\end{table}

\begin{table}[ht]
\centering
\resizebox{\columnwidth}{!}{%
\begin{tabular}{l|cccccc}
\toprule
\multirow{2}{*}{\textbf{Modules}} & \multicolumn{6}{c}{\textbf{Runtime (ms)}} \\
 & t=0 & t=0.1 & t=0.2 & t=0.3 & t=0.4 & t=0.5 \\
\midrule
Global Feature Extractor & 65.0 & 58.6 & 52.7 & 50.2 & 48.6 & 47.9 \\
Global Retrieval & 0.1 & 0.1 & 0.1 & 0.1 & 0.1 & 0.1 \\
Instance Segmentation & 38.4 & 38.8 & 38.6 & 38.7 & 38.6 & 38.5 \\
Receptive Field Expander & 7.9 & 3.2 & 1.8 & 1.3 & 1.0 & 0.7 \\
Object Feature Extractor & 146.9 & 86.9 & 55.6 & 40.9 & 32.3 & 26.3 \\
Object-Aware Scoring & 2.9 & 2.2 & 1.7 & 1.5 & 1.4 & 1.2 \\
Fine-Grained Retrieval & 87.0 & 87.4 & 87.0 & 87.1 & 87.5 & 87.3 \\
\midrule
\textbf{Total} & \textbf{349.5} & \textbf{278.6} & \textbf{238.8} & \textbf{221.1} & \textbf{210.9} & \textbf{203.4} \\
\bottomrule
\end{tabular}%
}
\caption{Mask R-CNN \& DINOv2 Runtime.}
\label{tab:module_runtime_md}
\end{table}

\begin{table}[ht]
\centering
\resizebox{\columnwidth}{!}{%
\begin{tabular}{l|cccccc}
\toprule
\multirow{2}{*}{\textbf{Methods}} & \multicolumn{6}{c}{\textbf{Accuracy (\%)}} \\
 & t=0 & t=0.1 & t=0.2 & t=0.3 & t=0.4 & t=0.5 \\
\midrule
AirRoom-MaskRCNN-ResNet & 92.70 & 92.68 & 92.58 & 92.59 & 92.22 & 92.15 \\
AirRoom-MaskRCNN-DINOv2 & 87.67 & 87.62 & 87.10 & 87.20 & 87.24 & 87.09 \\
\bottomrule
\end{tabular}%
}
\caption{Mask R-CNN \& ResNet / DINOv2 Accuracy.}
\label{tab:module_accuracy}
\end{table}

\begin{figure}[ht]
    \centering
    \includegraphics[width=\columnwidth]{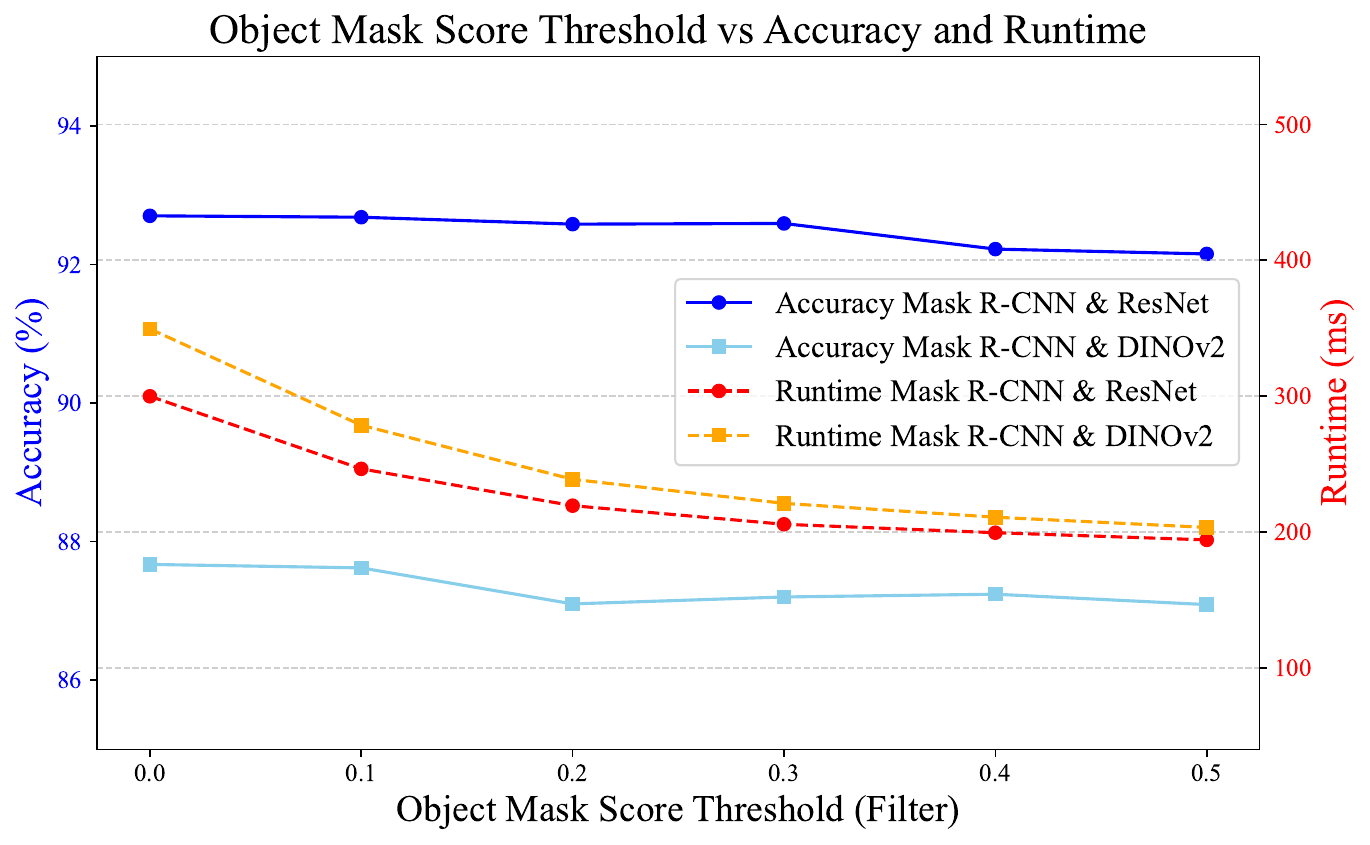}
    \caption{As the object mask score threshold increases, AirRoom's performance experiences a slight decline; however, the efficiency improves significantly.}
    \label{fig:runtime}
\end{figure}

\begin{table}[ht]
\centering
\begin{tabular}{l|cc}
\toprule
\multirow{2}{*}{\textbf{Modules}} & \multicolumn{2}{c}{\textbf{Runtime (ms)}} \\
 & ResNet & DINOv2 \\
\midrule
Global Feature Extractor & 42.5 & 56.2 \\
Global Retrieval & 0.1 & 0.1 \\
Instance Segmentation & 352.6 & 343.2 \\
Receptive Field Expander & 0.7 & 0.6 \\
Object Feature Extractor & 51.1 & 66.6 \\
Object-Aware Scoring & 2.2 & 2.1 \\
Fine-Grained Retrieval & 87.8 & 87.4 \\
\midrule
\textbf{Total} & \textbf{538.5} & \textbf{557.6} \\
\bottomrule
\end{tabular}%
\caption{Semantic-SAM \& ResNet / DINOv2 Runtime.}
\label{tab:module_runtime_ssam}
\end{table}

\begin{table}[ht]
\centering
\begin{tabular}{l|c|c}
\toprule
\textbf{Methods} & \textbf{Runtime (ms)} & \textbf{Accuracy (\%)}\\
\midrule
CVNet & 111.3 & 11.71 \\
DINOv2 & \textbf{16.7} & 53.91 \\
Patch-NetVLAD & 100.5 & 64.86 \\
AnyLoc & 45.5 & 89.69 \\
\rowcolor{Lavender} AirRoom & 194.2 & \textbf{92.15} \\
\bottomrule
\end{tabular}
\caption{Runtime Comparison with State-of-the-Art Methods.}
\label{tab:runtime_comparison}
\end{table}

When Mask R-CNN is used for instance segmentation, \tref{tab:module_runtime_mr} demonstrates that increasing the object mask score threshold significantly reduces the runtime of the Object Feature Extractor when ResNet is employed. This is attributed to the reduced number of objects and patches requiring processing. A similar trend is observed with DINOv2 as the Object Feature Extractor, as shown in \tref{tab:module_runtime_md}. Additionally, \tref{tab:module_accuracy} indicates that AirRoom's performance remains largely unaffected by the rise in the object mask score threshold, regardless of the chosen Object Feature Extractor. This observation is further illustrated in \fref{fig:runtime}. However, when Semantic-SAM is used for instance segmentation, AirRoom faces efficiency challenges due to Semantic-SAM's significantly slower performance, as detailed in \tref{tab:module_runtime_ssam}.

\tref{tab:runtime_comparison} compares runtime across methods. AirRoom requires 80ms more than CVNet but achieves over 80\% performance improvement. Compared to Patch-NetVLAD, AirRoom's runtime is approximately double, with a performance gain exceeding 30\%. While DINOv2 completes tasks in 10–20ms, AirRoom adds 170ms and improves performance by over 40\%. Relative to AnyLoc, AirRoom increases runtime by just over 150ms but captures an additional 20\% of the remaining performance potential. These results demonstrate that AirRoom delivers significant performance gains even within limited improvement margins, underscoring its effectiveness despite incremental runtime.

Currently, AirRoom allocates approximately 90ms to Fine-Grained Retrieval, utilizing LightGlue for feature matching. Exploring more lightweight and faster alternatives could further enhance efficiency. In real-world applications such as Real-Time Navigation, room reidentification times between 50–200ms are generally acceptable, with accuracy as the primary concern. While AirRoom is slightly slower than some baselines, it achieves substantial accuracy improvements, effectively balancing runtime and performance. This makes AirRoom well-suited for practical scenarios, meeting real-world runtime requirements while maintaining high reliability and precision.

\end{document}